# Container damage detection using advanced computer vision model – Yolov12 vs Yolov11 vs RF-DETR – A comparative analysis

Subhadip, Kumar

Canadian Pacific Kansas City, skuma19@wgu.edu

**ABSTRACT:** Containers are an integral part of the logistics industry and act as a barrier for cargo. A typical service life for a container is more than 20 years. However, overtime containers suffer various types of damage due to the mechanical as well as natural factors. A damaged container is a safety hazard for the employees handling it and a liability for the logistic company. Therefore, a timely inspection and detection of the damaged container is a key for prolonging service life as well as avoiding safety hazards.

In this paper we will compare the performance of the damage detection by three state of the art advanced computer vision model – Yolov12, Yolov11 and RF-DETR. We will use a dataset of 278 annotated images to train, validate and test the model. We will compare the mAP and precision of the model. The objective of this paper is to identify the model that is best suited for container damage detection.

The result is mixed. mAP@50 score of Yolov11 and 12 was 81.9% compared to RF-DETR which was 77.7%. However, while testing the model for not so common damaged containers RF-DETR model outperformed the others overall, exhibiting superiority to accurately detect both damaged containers as well as damage occurrences with high confidence.

## 1 INTRODUCTION

Shipping containers became a vital component of logistics as well as freight railroad industry and a key component of global trade operation. Shipping containers undergo several stresses everyday – that includes improper handling, weather condition – storm, flood, earthquake, extreme heat and cold, fire and collision. Common types of damage include denting, scratching, bending. Such damage usually occurs due to improper handling, overloading, fire or prolonged expose to extreme weather conditions. A damaged container is a security hazard as well as a liability. Damaged containers pose serious risks to workers, cargo and transport infrastructure. A compromised structure may collapse, leak and cause accidents while handling. Moreover, several international shipping standards require that containers meet specific safety requirements and quality benchmarks.

Therefore, containers inspection is crucial. It prevents damage to the expensive cargo, preserves the company's reputation and trust and most importantly prevents any safety hazards. Early detection and fixing damaged cargo are much more cost effective than the liabilities incurred by the damaged cargo. A manual inspection is costly and error prone. With the advancement of computer vision model damaged container detection is simplified and automated.

A previous study of container damage detection was conducted using Yolo-NAS by Phoung et al. [11] achieved a high mAP value of 91.2%. They also concluded that Yolo-NAS model outperformed Roboflow 3.0. In 2019, Valente et al. performed a study to detect waste containers using Yolo models [12]. Tang et al. performed another study at Tianjin port for damage detection [13] in 2020. Since September 2024 three new CNN based model evolved – Yolov11, Yolov12 and RF-DETR. We will train, validate and test damaged container dataset containing 278 annotated images. We will then compare the performance of the model. Finally, we will try this model against 3 damaged container images and measure the efficiency of all three models.

The following research questions guide the focus of this thesis:

RQ1: Which of the models YOLOv11, YOLOv12 and RF-DETR performs better in accurately detecting and locating damaged containers?

RQ2: Can deep learning models like YOLO be an effective and practical alternative to manual and traditional methods in detecting damaged containers?

## 2 RELATED WORK

As per containerstatistic.com around 187 million TEUs (twenty-foot equivalent units) are moved annually, which breaks down to 512,000 containers per day containing a staggering flow of goods, from electronics to coffee beans. Containers are essential links in the global logistics chain, seamlessly transferring goods between ships, rail, and trucks in intermodal operations. Approximately 25% of containers undergo some or other kinds of damage such as dents, cracks or structural failures. When a container is damaged—whether due to impact, corrosion, or structural fatigue—it can trigger cascading disruptions across the entire supply network. At ports, a damaged container may stall crane operations, require manual inspection, or necessitate re-routing of cargo, leading to delays and additional labor costs. On freight railroads, compromised containers raise safety concerns, especially if they contain hazardous materials, and may force carriers to remove them from service, causing scheduling setbacks. In intermodal systems, where speed and coordination are paramount, damaged containers disrupt the handoff between transport modes, risk cargo loss or contamination, and increase insurance claims. The cumulative impact of damaged containers not only affects operational efficiency but also diminishes customer trust and increases environmental waste due to damaged goods. Robust inspection and damage detection systems are thus critical to maintaining the integrity and reliability of intermodal transport. Fig 1 illustrates different types of container damage. Despite being the most used approach for assessing container damage, manual inspections are fundamentally limited by their dependence on human labor, susceptibility to errors, and slow execution. As global trade volumes continue to surge, these shortcomings have become increasingly detrimental—creating operational delays, reducing throughput efficiency, and straining resources at busy ports and intermodal terminals. The reliance on subjective judgment also introduces variability in damage detection, further complicating logistics planning and quality control in fast-paced, high-volume environments.

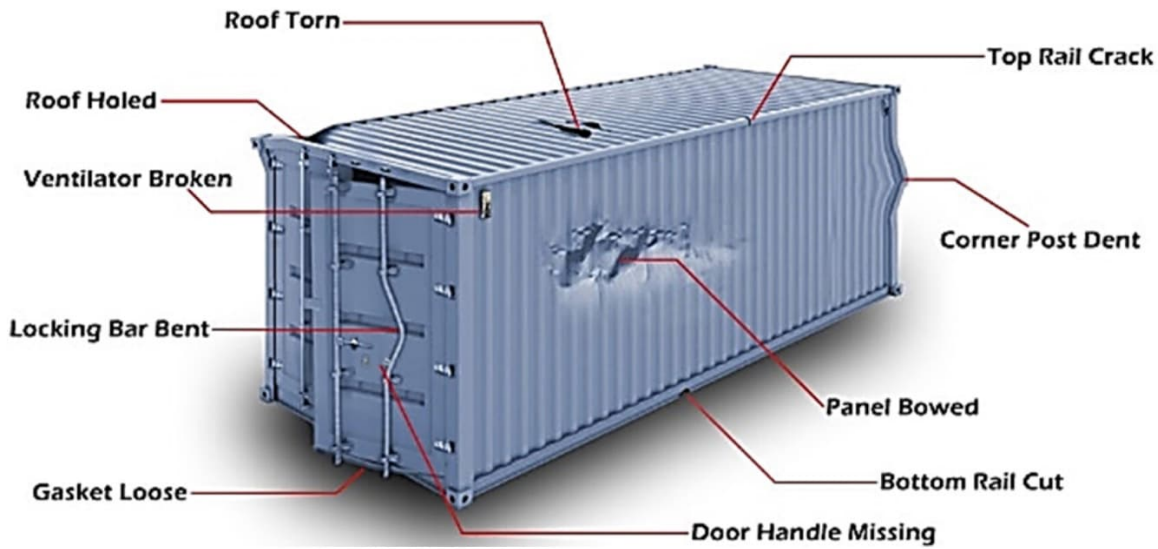

Figure 1: Type of container damage [14]

Fig 2 illustrates different sections of containers. A standard intermodal shipping container is a marvel of modular engineering, designed for durability and efficient cargo handling. Some key structural sections are front and rear end walls, side walls, roof panel, floor assembly, door assembly, corner castings and posts.

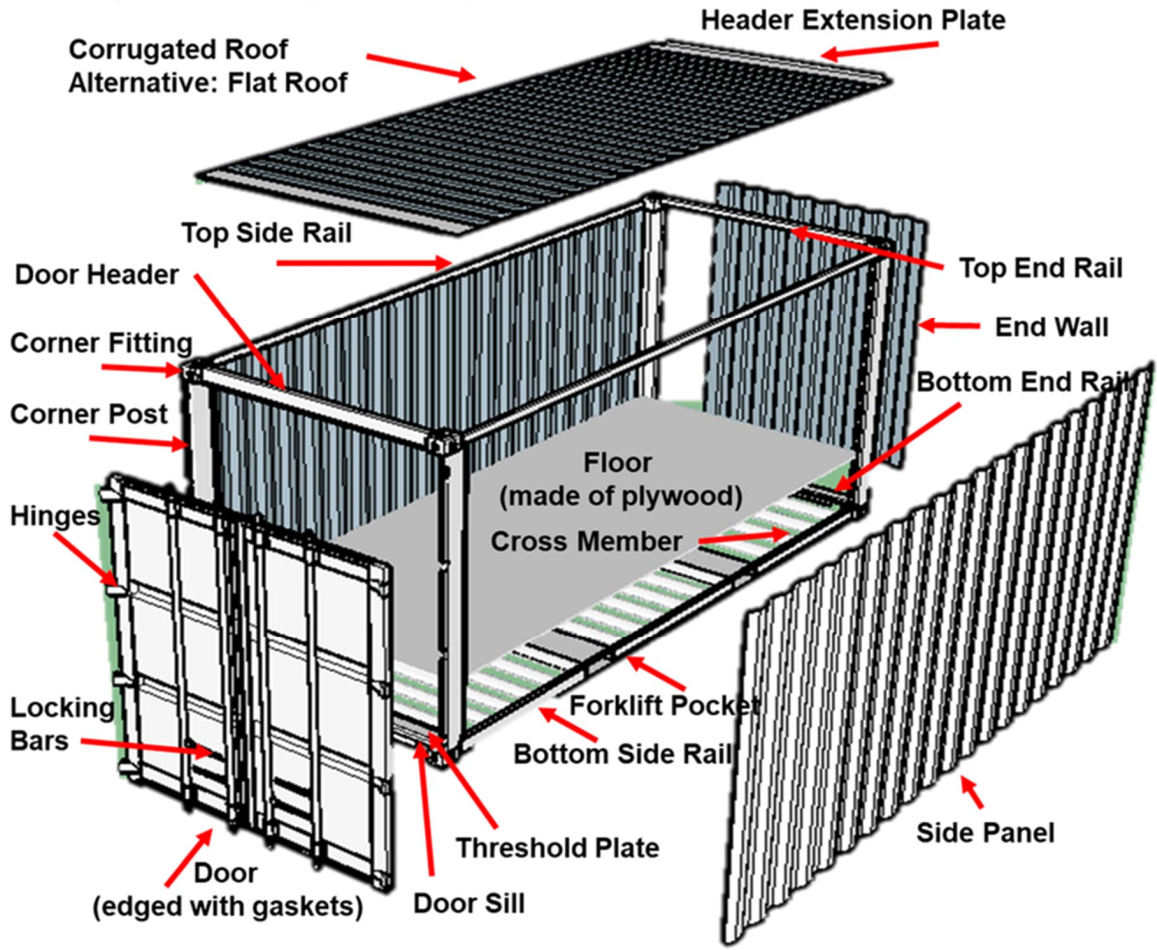

Figure 2: Different sections of container [15]

In the next section we will discuss the basic architecture of three new single-shot object detector computer vision models based on Convolutional Neural Network (CNN).

## 2.1 Key Features of Yolov11

You Only Look Once version 11 model was released around September 2024. This version offered significant improvements in speed, accuracy and feature extraction. Any Yolo model consists of three key components – Backbone, Neck and Head. Backbone acts as a feature extractor, Neck acts as a bridging stage, finally the head module performs the prediction tasks, producing output for object detection and classification based on advanced features.

While retaining the Spatial Pyramid Pooling – Fast (SPPF) block in this Yolo release backbone, v11 introduced a new cross stage partial and spatial attention (C2PSA) block as it enhances spatial attention in the feature maps. SPPF and C2PSA block extract multi-scale features from the input image efficiently as well as accurately.

A significant change in the Neck section in v11 is C2f block was replaced by C3k2 block. Along with other convolution layers and upsampling blocks neck aggregate features and pass it to the head block.

The C3k2 block was also introduced in the head section where it processes the feature maps that came from the neck section. C3k2 blocks offer faster processing and parameter efficiency. The head also introduced the CBS (Convolution-BatchNorm-Silu) layer which helps to extract relevant features, improve model performance by utilizing Sigmoid Linear Unit (SiLU) activation. Finally, Conv2D layers reduce the number of outputs for bounding box coordinates and class predictions.

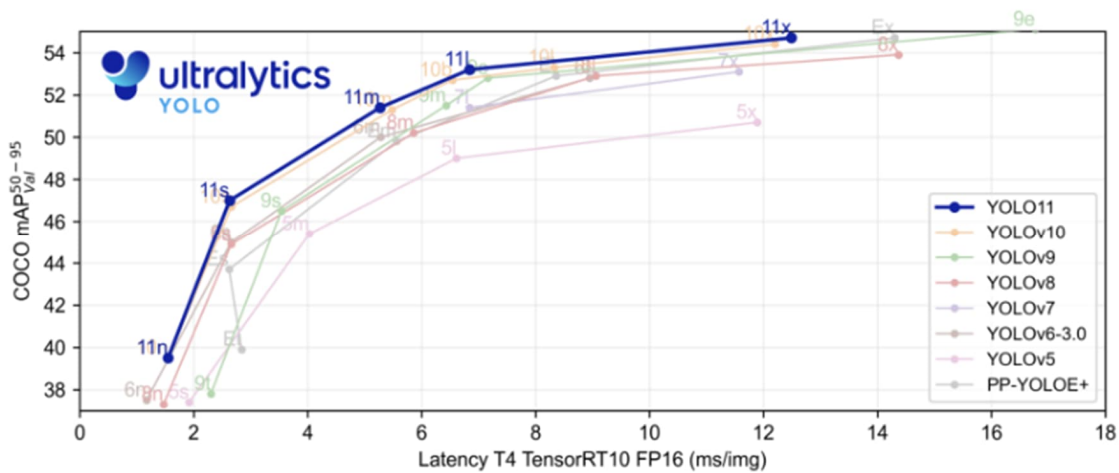

Figure 3: Yolov11 Benchmarking [3]

## 2.2 Key Features of Yolov12

Yolov12 introduced an attention-centric architecture that is totally different from traditional CNN-based architecture [5] in February 2025. Two major improvement factors for Yolov12 – Area Attention and Residual Efficient Layer Aggregation Networks (R-ELAN).

Area Attention: Unlike traditional attention mechanisms that attend to individual elements such as a word in a sentence (one-dimensional) or pixels in an image (two-dimensional). An area is a collection of structurally adjacent items in the memory.

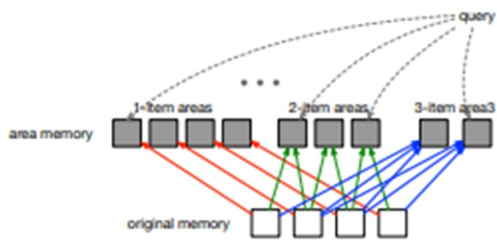

Figure 4: One Dimensional area attention [7]

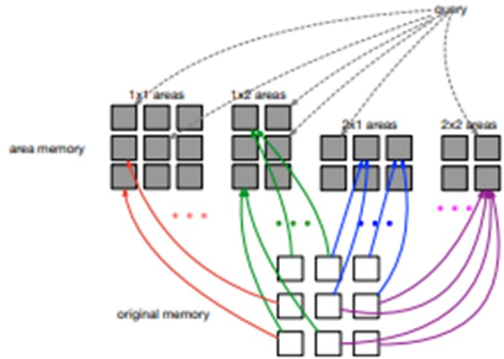

Figure 5: Two-Dimensional area attention [7]

Area attention allows models to focus on groups of adjacent elements as seen in Fig 4 and 5 both one-dimensional and two-dimensional areas. The GitHub repo [6] provides a pyTorch implementation of area attention mechanism for context understanding in tasks such as image processing and natural language processing. The key features are single-head and multi-head implementation, integration with existing architectures and unit tests and setup.

Residual Efficient Layer Aggregation Networks (R-ELAN): In previous Yolo models both CSPNet (Cross-Stage Partial Network) and ELAN (Efficient Layer Aggregation Network) are designed to improve deep learning architectures. While CSPNet focuses on reducing computational complexity by splitting feature maps and processing them separately before merging, ELAN optimizes gradient flow by controlling the shortest and longest paths, allowing deeper networks to converge more effectively.

The Efficient Layer Aggregation Network (ELAN) is designed to optimize feature extraction and aggregation in deep learning models, particularly in object detection. When constructing an efficient network, designers typically focus on optimizing a limited set of parameters, including computational requirements and density. The architecture of YOLOv7 is built upon Efficient Layer Aggregation Networks (ELAN), which strategically manages both the shortest and longest gradient paths. This approach ensures that deeper networks can effectively converge and enhance learning capability.

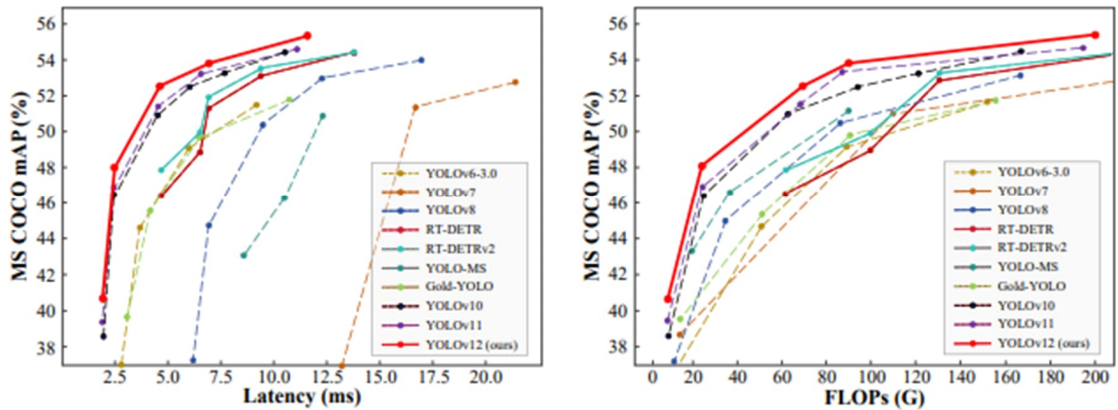

Figure 6: Benchmarking Yolov12 compared to previous versions: mAP vs. Latency and mAP vs. FLOPs

## 2.3 Key Features of RF-DETR

RF-DETR or RoboFlow Detection TRansformer model released in March 2025. Instead of focusing on region proposals and anchor boxes, DETR uses a transformer encoder-decoder to process image features and predict object locations and classes. It removes the need for complex components like non-maximum suppressions (NMS) and anchor generation and makes the architecture cleaner. With minor modifications, DETR can also perform panoptic segmentation, unifying object detection and segmentation in a single model [10]. DETR was introduced by Facebook AI Research in 2020 and RF-DETR model developed by Roboflow combines the innovations of Deformable DETR and LW-DETR and incorporates the DINOv2 backbone to improve background modeling and domain matching [8]. There were some initial shortcomings of DETR model such as slow convergence and high computational demands but new versions of DETR. RF-DETR model achieves 60.5 mAP at 25 FPS on an NVIDIA T4 [4]. According to Sapkota et al. RF-DETR is the first model that surpass mAP@60% on the MS COCO dataset. Figure 7 displays the performance of RF-DETR model.

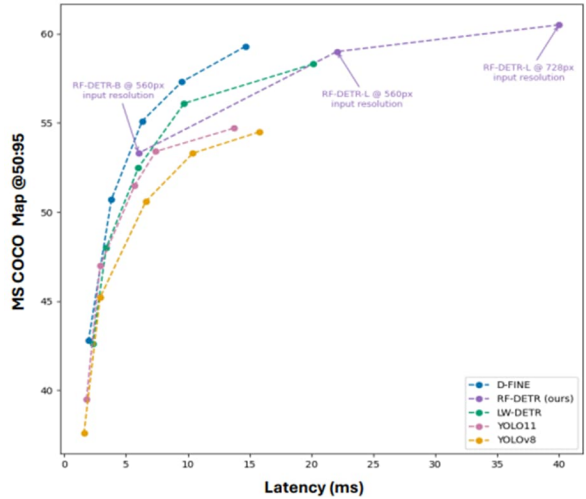

Figure 7: Benchmarking RF-DETR compared to D-FINE, LW-DETR Yolov8, Yolov11 [8]

## 3 DATASET

The dataset used for this study contains 278 annotated images. The images are divided into 234(84%), 25(9%) and 19(7%) for training, validation and testing. The dataset is created from publicly available images for damaged containers in the Roboflow app and then mixed with other publicly available images.

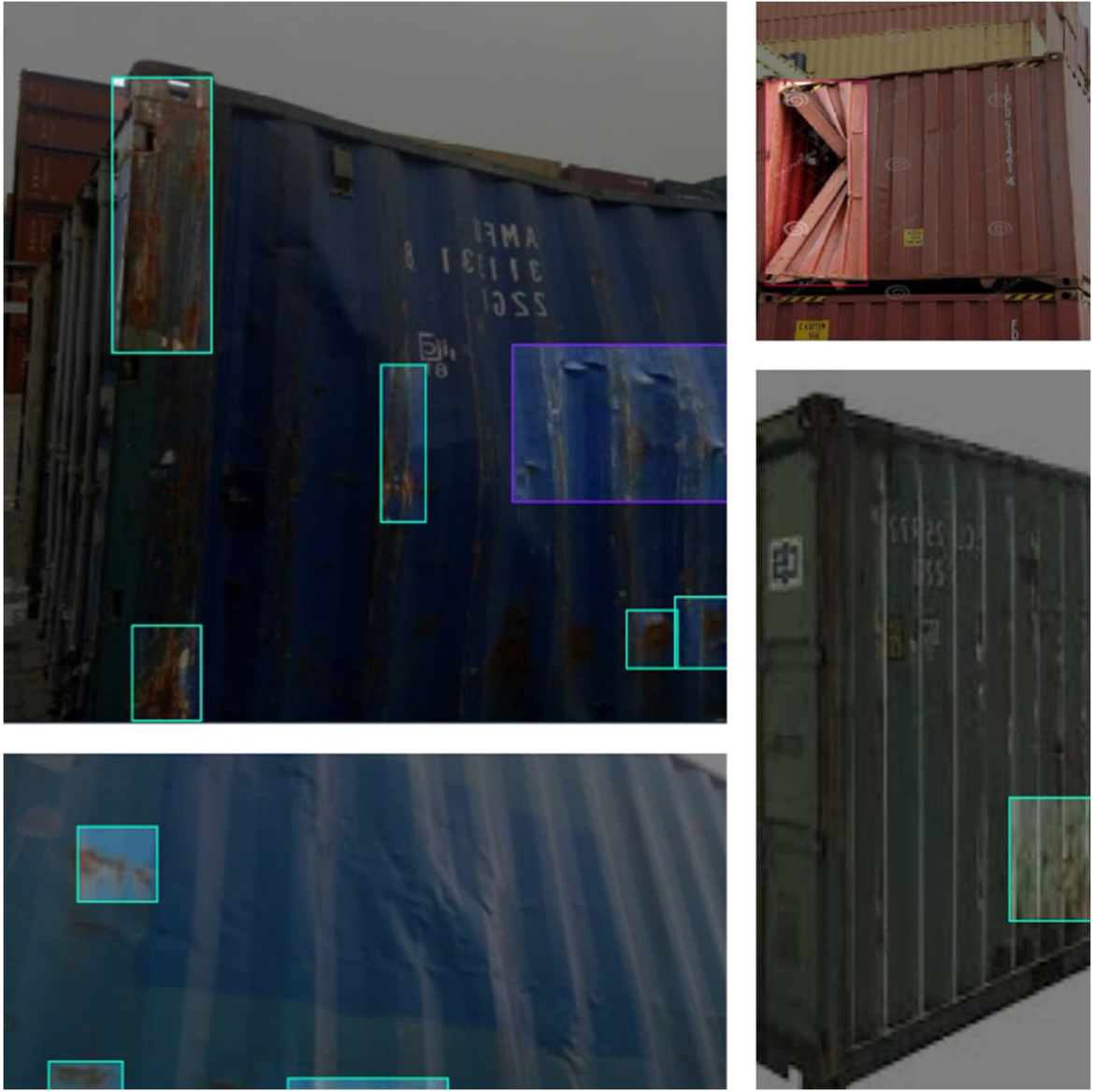

Figure 8: Images used for the dataset

## 4 COMMON METRICS

Here are the metrics that we are going to use to assess the model's performance.

### 4.1 Precision

In object detection, precision measures how accurate your model's positive predictions are, essentially, how many of the objects it said it found were correct.

$$\text{Precision} = \frac{\text{True Positives}}{\text{True Positives + False Positives}}$$

True Positives (TP): Correctly detected objects.
False Positives (FP): Incorrect detections, the model thought were objects but weren't correct objects
So, more FP means lower precision.

### 4.2 Recall

In object detection, recall measures how well a model finds *all* the actual objects in an image—it's about completeness.

$$\text{Recall} = \frac{\text{True Positives}}{\text{True Positives + False Negatives}}$$

True Positives (TP): Objects correctly detected.
False Negatives (FN): Objects that were present but missed by the model.

High recall means the model rarely misses objects—it's good at finding everything it's supposed to.

### 4.3 IoU

IoU, or Intersection over Union, is a key metric used in object detection and image segmentation to evaluate how well a predicted bounding box matches the ground truth box.

$$\text{IoU} = \frac{\text{Area of Overlap}}{\text{Area of Union}}$$

Area of Overlap: The region where the predicted and ground truth boxes intersect.
Area of Union: The total area covered by both boxes combined.

### 4.4 mAP

mAP, or mean Average Precision, is a widely used metric for evaluating the performance of object detection models. It tells you how well a model is at both detecting and correctly classifying objects in images.

$$\text{mAP@0.5} = \frac{1}{N} \sum_{i=0}^{N} (AP_i)$$
$$\text{where IoU} \geq 0.5$$

## 5 RESULTS

The result of the performance model is mixed in nature. Yolov11 and Yolov12 model demonstrated similar mAP@50% and similar Recall however precision of Yolov11 surpass the other two. RF-DETR performed well for precision compared to Yolov12 but fell short in mAP and recall.

| Model | Metrics |
|---|---|

|  | mAP@50 | Precision | Recall |
|---|---|---|---|
| Yolov11 | 81.90% | 91.90% | 76.10% |
| Yolov12 | 81.90% | 78.90% | 78.40% |
| RF-DETR | 77.70% | 83.80% | 73% |

Table 1: Model Performance

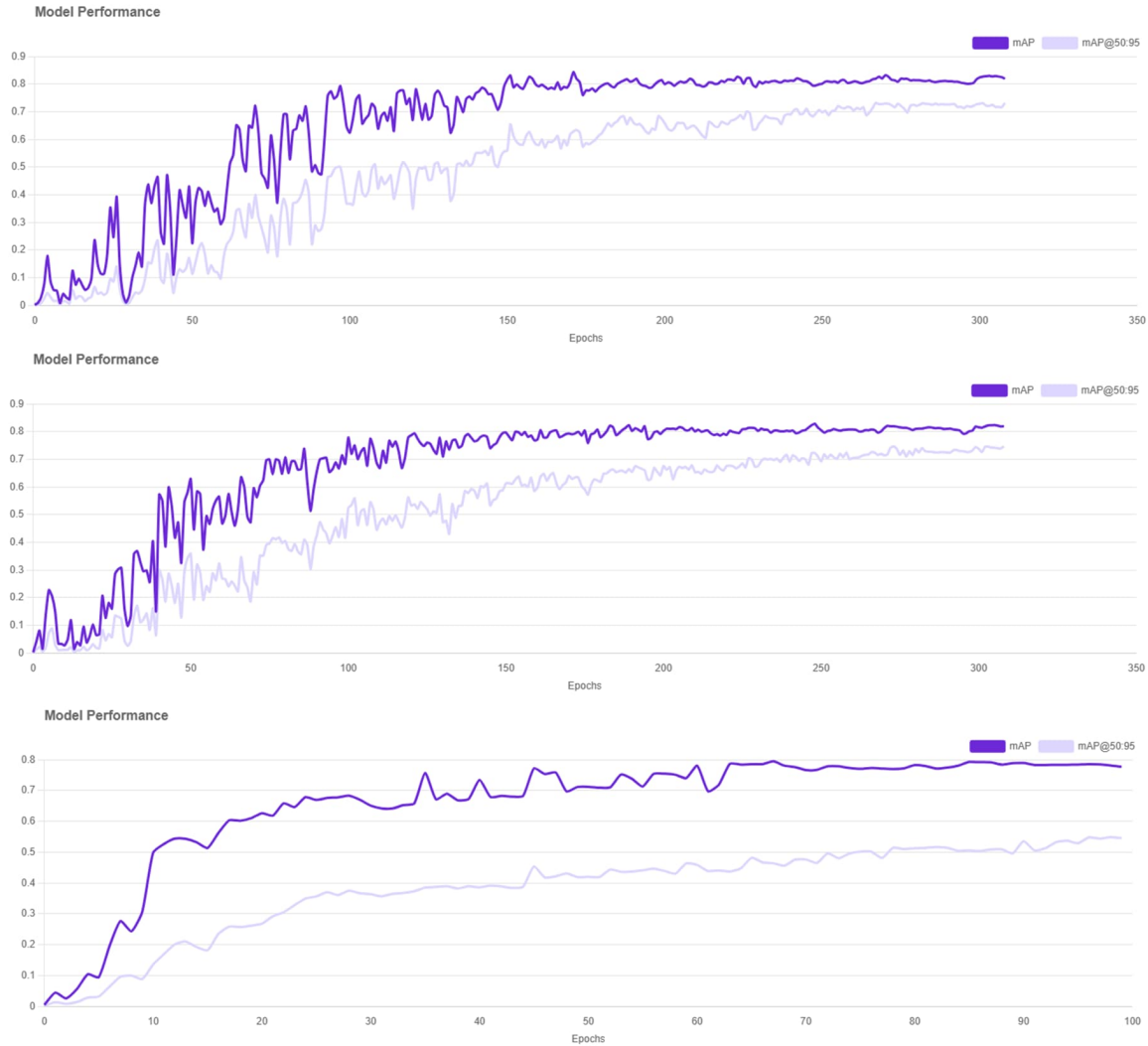

Figure 9: How mAP changes as the model trains over time (Epochs) – Yolov11, Yolov12 and RF-DETR

All three models then tested against other images (not part of original dataset) and this time result varied significantly.

Test1: A fully fire damaged container [Fig 10] was tested against all three models and a 50% confidence threshold. Yolov11 model is unable to determine any damage section, Yolov12 model is only able to determine a portion of side wall with 51% confidence however RF-DETR model able to determine all the sections (front and side wall) with more than 90% confidence.

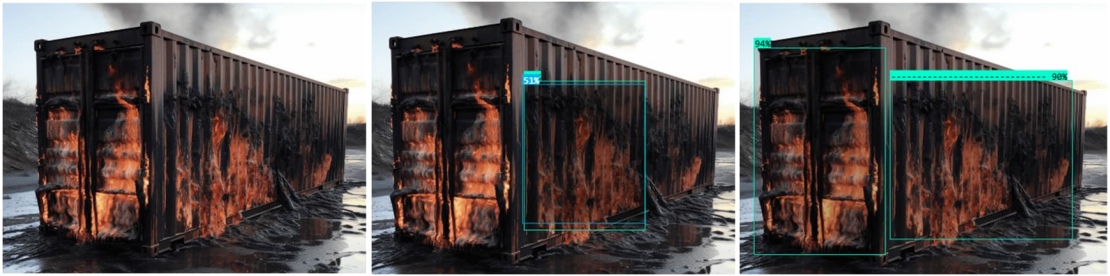
Figure 10: Fire damaged container for Test1 – Yolov11, Yolov12 and RF-DETR

Test 2: Another two stacked fire-damaged containers tested against similar conditions. Yolov11 model able to predict damage on side wall of one container with 89% confidence, Yolov12 model is unable to determine any damage, RF-DETR model can predict one section with 93% confidence [Fig 11].

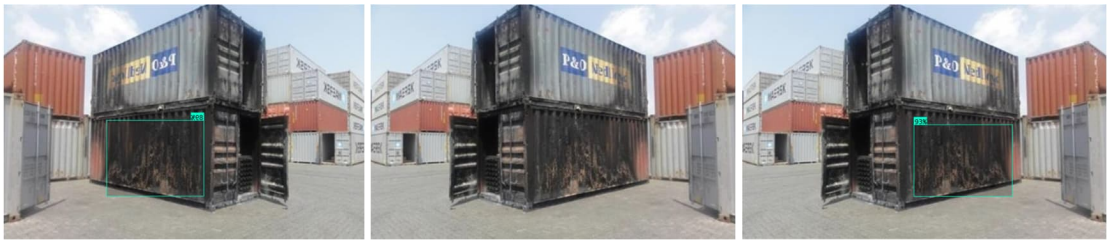
Figure 11: Fire damaged container for Test2 – Yolov11, Yolov12 and RF-DETR

Test 3: A fully bend container was tested against all three models and a 50% confidence threshold. Yolov11 model was able to identify two damaged sections with 59% and 69% confidence, Yolov12 model was able to identify one section with 87% confidence level, finally RF-DETR model able to identify 3 different sections with 54%, 55% and 81% confidence level [Fig 12].

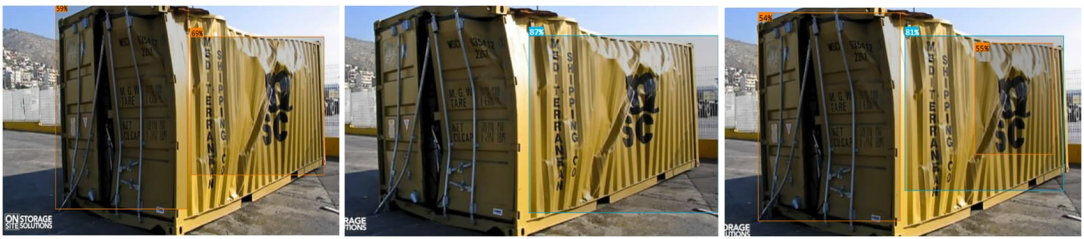
Figure 12: Bend container for Test 3 – Yolov11, Yolov12 and RF-DETR model
From the result we will try to answer our research questions.

Which of the models YOLOv11, YOLOv12 and RF-DETR performs better in accurately detecting and locating damaged containers?

All three models performed well and have mixed results – Yolov11 and Yolov12 model performed well over RF-DETR with high mAP value. While the model further tested beyond the original dataset, RF-DETR model surpassed the other two models. Therefore, there is no clear winner, but all three models displayed showed strong performance.

Can deep learning models like YOLO be an effective and practical alternative to manual and traditional methods in detecting damaged containers?

Yes, all three deep learning models showed strong performance and can be effectively used as an alternative to manual methods to detect damaged containers.

## 6 CONCLUSION

This study presented a comprehensive comparative analysis of three state-of-the-art object detection models—YOLOv12, YOLOv11, and RF-DETR to detect damaged container. The result highlights strong performance for all three models. RF-DETR excelled in complex localization scenarios due to its region-aware transformer attention. These findings provide valuable guidance for selecting and deploying object detection models in industrial damage assessment workflows, where accuracy, speed, and adaptability are critical. Future work may explore hybrid approaches or fine-tuning strategies to further enhance model robustness under variable conditions.

## 7 ACKNOWLEDGMENTS

I would like to thank anonymous reviewers for the comments and suggestions.